\pdfoutput=1

\documentclass[11pt]{article}

\usepackage[preprint]{acl}

\usepackage{times}
\usepackage{latexsym}

\usepackage[T1]{fontenc}

\usepackage[utf8]{inputenc}

\usepackage{microtype}

\usepackage{inconsolata}

\usepackage{graphicx}

\usepackage{booktabs}
\usepackage{listings}
\usepackage{float}
\usepackage{caption}
\usepackage{subcaption}
\usepackage{multirow}

\title{LocateBench: Evaluating the Locating Ability of Vision Language Models}

\author{Ting-Rui Chiang ~\;~ Joshua Robinson ~\;~ Xinyan Velocity Yu ~\;~ Dani Yogatama \\
         \\ University of Southern California \\ \texttt{\{tingruic,joshua.j.robinson,xinyany,yogatama\}@usc.edu}
         }

\begin{document}
\maketitle
\begin{abstract}
The ability to locate an object in an image according to natural language instructions is crucial for many real-world applications. In this work we propose LocateBench, a high-quality benchmark dedicated to evaluating this ability. We experiment with multiple prompting approaches, and measure the accuracy of several large vision language models. We find that even the accuracy of the strongest model, GPT-4o, lags behind human accuracy by more than 10\%.
\footnote{We release the dataset at \url{https://usc-tamagotchi.github.io/locate-bench/}.}

\end{abstract}

\section{Introduction}

Locating an object in an image is an essential part of many real-world tasks. For example, in web page navigation tasks~\cite{mind2web,webshop,zhou2023webarena}, the agent needs to locate buttons or other HTML elements before deciding the next action to take, and in robotics tasks~\cite{alfred,NEURIPS2021_021bbc7e,pmlr-v205-li23a}, the agent needs to locate a specific object based on the grounded input. 
This ability also contributes to many downstream tasks such as visual question answering and image captioning.
Despite numerous studies of the performance of vision language models (VLMs) on these downstream tasks \citep{liu2023visual,li-etal-2023-evaluating,zhang2023m3exam,2023GPT4VisionSC,yu2024mm,zhu2024minigpt}, there is no direct measurement of the locating ability of VLMs, an upstream ability that greatly affects downstream task performance.

\begin{figure}[t]
    \centering
    \begin{subfigure}[t]{0.23\textwidth}
            \centering
            \includegraphics[width=\linewidth]{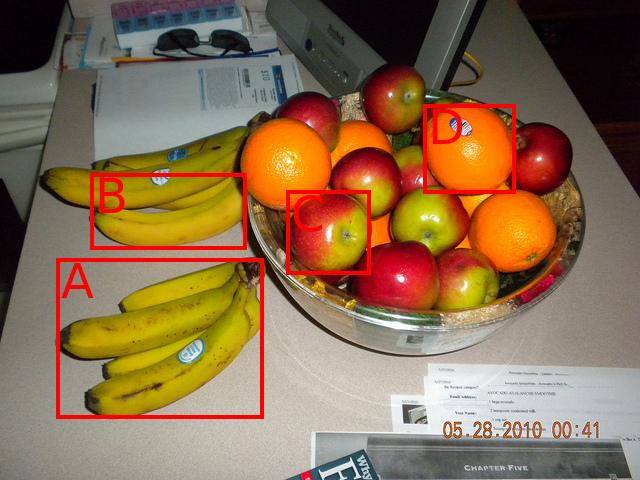}
            \caption{Which one contains the bunch of bananas that has only one sticker?}
            \label{fig:ex-fine-grained}
    \end{subfigure}
    \hspace{0.01\textwidth}
    \begin{subfigure}[t]{0.23\textwidth}
            \centering
            \includegraphics[width=\linewidth]{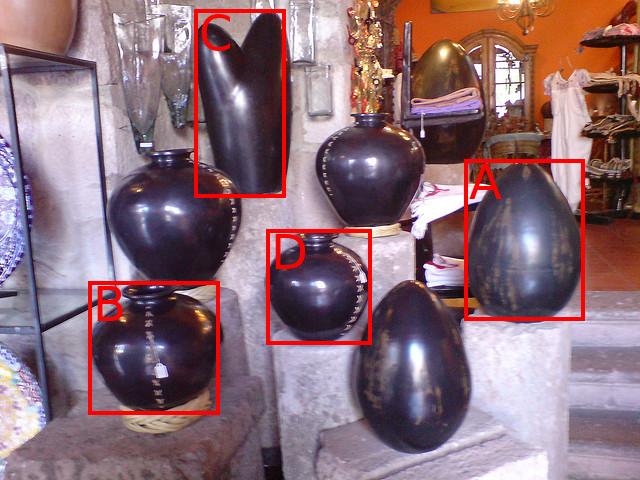}
            \caption{Which one contains the tallest oval-shaped vase?}
            \label{fig:ex-size}
    \end{subfigure}
    \begin{subfigure}[t]{0.23\textwidth}
            \centering
            \includegraphics[width=\linewidth]{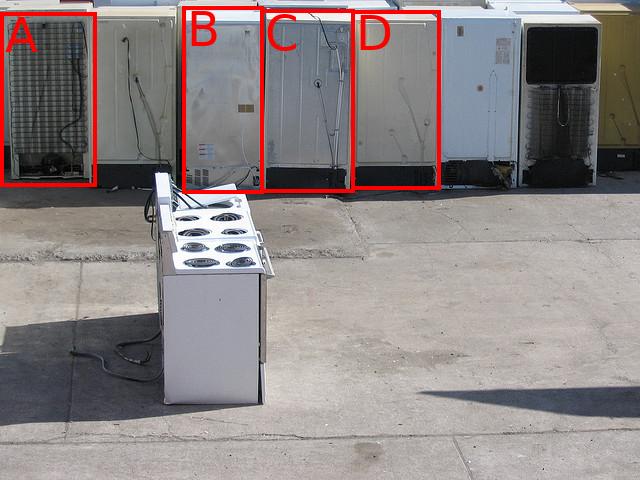}
            \caption{Which one contains the third fridge counting from the left?}
            \label{fig:ex-counting}
    \end{subfigure}
    \hspace{0.01\textwidth}
    \begin{subfigure}[t]{0.23\textwidth}
            \centering
            \includegraphics[width=\linewidth]{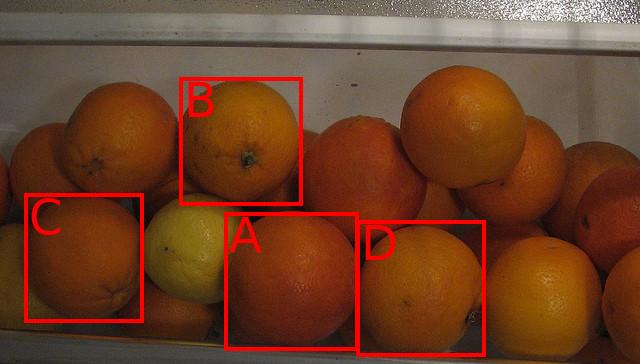}
            \caption{Which one contains the fruit that is to the left of the yellow one?}
            \label{fig:ex-rel}
    \end{subfigure}
    \caption{
    Some examples from LocateBench.
    Questions in our dataset can be categorized into fine-grained descriptions (\ref{fig:ex-fine-grained}), relative size (\ref{fig:ex-size}), counting (\ref{fig:ex-counting}) or relative location (\ref{fig:ex-rel}). 
    }
    \label{fig:example}    
\end{figure}


To address this, we propose LocateBench, a benchmark that requires VLMs to select the correct bounding box out of four candidate boxes in each image based on natural language questions in English (Figure~\ref{fig:example}).
The multiple choice setup of LocateBench allows for evaluation of VLMs that do not have a dedicated input/output field for bounding boxes or image segmentation masks.


To our knowledge, our dataset is the first expert-annotated, high-quality benchmark designed for evaluating the locating ability of VLMs.
Most previous datasets do not specifically focus on locating objects in images (more discussion in \S{\ref{sec:related-datasets}}).
An existing dataset, Pointing QA~\cite{zhu2016cvpr}, shares the same task formulation as ours, yet the candidate objects in the images it employs tend to be either too small or overlapping with each other.
LocateBench, in comparison, has less ambiguity and noise while also having higher complexity (\S{\ref{sec:pointing-qa}}).

\section{LocateBench}

\subsection{Dataset Formulation}

LocateBench is a multiple choice question dataset. Each sample includes an image, a description formulated as a question, and four bounding boxes representing candidate answers to the question. VLMs are tasked with choosing the bounding box that best answers the question.
 

\subsection{Dataset Construction}
\label{sec:construct}

LocateBench is constructed based on the RefCOCO series datasets.
These datasets contain descriptions of objects in images from the COCO dataset~\cite{mscoco}.
We utilize these descriptions to construct our LocateBench dataset.
Through manual inspection, we find the descriptions in RefCOCO-g~\citep{refcocog} are more specific and detailed than their counterparts in RefCOCO~\citep{refcoco} or RefCOCO+~\citep{refcocop}. Therefore, we prioritize using the object descriptions in RefCOCO-g where possible.

We construct LocateBench with the following steps:
\begin{enumerate}
    \item 
    We discard objects with no descriptions found in the RefCOCO series dataset.
    \item 
    Based on the super-category and bounding box information provided in the COCO dataset, we filter the COCO dataset and keep only the images that contain at least four objects in the same category.
    To ensure that there is no ambiguity, we only consider the sets of objects whose bounding boxes do not significantly overlap with each other.
    In particular, we ensure that the width and height of the overlapping area are no more than 10 pixels each.
    We further discard objects whose size in the image is too small (i.e., objects whose bounding box width or height is less than 75 pixels). 
    We have 1317 examples after filtering.
    This size is comparable with the test set size of GSM8k~\cite{cobbe2021training}, which is a commonly used benchmark dataset for math reasoning capability.
    \item 
    Next, two authors \textit{manually} inspect and edit the descriptions.
    From our inspection, we find that in RefCOCO and RefCOCO+, the descriptions are sometimes not specific enough to locate the target object or are oversimplified.
    For example, the descriptions sometimes refer a man in a blue top as ``blue man'' or uses the relative location of items to the observer such as ``4 pm'' (meaning the item is located in the 4 o'clock position).
    The crowdworkers of RefCOCO and RefCOCO+ may have chosen to write descriptions in this way to save time.
    However, this might cause unnecessary ambiguity.
    Therefore, to ensure the quality of our benchmark, we re-annotate the descriptions without using crowdworkers.
    We re-annotate 683 examples in total.
    \item 
    We then use the LLM Reka-Core~\cite{ormazabal2024reka} to convert the descriptions to fluent English which-questions, e.g., ``Which one contains the tallest oval-shaped vase?''
    We use seven demonstrations for in-context learning.
    \item
    Finally, we measure human performance by having two of the authors answer the collected questions, ensuring that these authors do not evaluate the questions that they inspected in the previous step. We observe human accuracy of 95\%.
    \item
    We re-inspect the examples where the two authors answered incorrectly.
    We edit the example description to ensure the authors agree on the answer.
\end{enumerate}

\section{Experiments on VLMs}

\begin{table*}[]
    \centering
    \begin{tabular}{llccccccc}
    \toprule
    \multirow{2}{*}{Dataset} & \multirow{2}{*}{Prompt} & \multicolumn{2}{c}{GPT} & \multicolumn{2}{c}{Gemini} & Claude-3 & \multicolumn{2}{c}{LLaVA-1.6}  \\
    & & 4o & 4T & 1.5p & 1.0p & Opus & Vicuna & Mistral \\
    \midrule
    \multirow{4}{*}{LocateBench} & ABCD       & \textbf{81.2} & 59.0 & 73.3 & 60.6 & 31.3 & 27.3 & 53.7 \\
                              & Colors     & \textbf{79.0} & 57.6 & 70.2 & 60.7 & 32.3 & 33.1 & 38.3 \\
                              & 1-by-1     & \textbf{60.4} & 48.7 & 41.8 & 41.7 & 21.8 & 25.6 & 26.5 \\
                              & Coordinate & \textbf{45.4} & 38.9 & 29.4 & 31.6 & 38.6 & 30.7 & 30.1 \\
    \midrule                                                    
                  Pointing QA & ABCD       & 78.2 & 66.7 & \textbf{79.7} & 61.6 & 29.7 & 25.4 & 55.8 \\
    \bottomrule
    \end{tabular}
    \caption{Model accuracy on LocateBench and Pointing QA~\cite{zhu2016cvpr} using the prompts in \S{\ref{sec:prompt-methods}}. 
    We use the ``pro'' versions of Gemini-1.0 and Gemini-1.5.}
    \label{tab:results}
\end{table*}

\subsection{Evaluating Methods}
\label{sec:prompt-methods}

To isolate the effect of prompt formats and precisely estimate the locating capability of LLMs, we prompt VLMs in the following formats:

\paragraph{Multi-choice by alphabet letters (ABCD)}
We draw the four candidate bounding boxes in red. Each box is assigned a letter (either A, B, C, or D), and this letter is placed in the top left corner of the box. We prompt the model with the template: 

\begin{table}[H]
    \centering
    \begin{tabular}{|p{0.9\linewidth}|}
    \hline
    \texttt{There are 4 bounding boxes (drawn in red rectangles) marked with A, B, C, D in the image. \{question\}}
\newline \newline
\texttt{Please answer in the following format:
Answer: (A|B|C|D)}
\\
\hline
    \end{tabular}
\end{table}
Here \texttt{\{question\}} is a placeholder for the which-question in our dataset (e.g, ``Which contains the tallest suitcase?'')

\paragraph{Multi-choice by colors}
We draw each bounding box in a different color (red, green, blue, or yellow), and prompt the model with the template:
\begin{table}[H]
    \centering
    \begin{tabular}{|p{0.9\linewidth}|}
    \hline
    \texttt{There are 4 bounding boxes drawn in color red, green, blue and yellow. \{question\}}
\newline \newline
\texttt{Please answer in the following format:}
\texttt{Answer: (red|green|blue|yellow)}
\\
\hline
    \end{tabular}
\end{table}

\paragraph{Multi-choice by coordinates}
Instead of drawing bounding boxes on the image, we provide the coordinates of the four candidates' bounding boxes in the prompt:
\begin{table}[H]
    \centering
    \begin{tabular}{|p{0.9\linewidth}|}
    \hline
    \texttt{There are 4 \{category\} in the image. Their bounding boxes (x, y, width, height) are \{b0\}, \{b1\}, \{b2\}, \{b3\} respectively. \{question\}}
\newline \newline
\texttt{Please output the bounding box in the following format:}

\texttt{Answer: (x, y, width, height)
}
\\
\hline
    \end{tabular}
\end{table}

In the template, \texttt{\{category\}} is the name of the super-category the candidates belong to. (Our dataset generation process ensures they belong to the same super-category.) \texttt{\{b1\}}, \texttt{\{b2\}}, \texttt{\{b3\}}, \texttt{\{b4\}} are the four candidates' bounding boxes following the format (x, y, width, height).

\paragraph{1-by-1}
For each question, we query the VLM multiple times.
Each time, we draw a red bounding box for a candidate and prompt the model with

\begin{table}[H]
    \centering
    \begin{tabular}{|p{0.9\linewidth}|}
    \hline
    \texttt{\{question\} Please output the answer in the following format:}

    \texttt{Answer: (Yes|No)}
\\
\hline
    \end{tabular}
\end{table}

Here \texttt{\{question\}} is a yes/no question, e.g., ``Does the red box contain the tallest suitcase?''. 
We choose the first candidate for which the model returns ``yes'' as the model's prediction.
If the model returns ``no'' for all the candidates, we pick the first candidate as its prediction.

\paragraph{Answer Extraction} We find that Claude Opus and the LLaVA models do not follow the format specified in our instructions.\footnote{Thus, for Claude and LLaVA, we remove the line for the answer format from the prompt.}
When we prompt GPT-4o/Turbo to output the coordinates, they do not always follow the format.
Therefore, instead of using a rule-based extraction method, we use Chat-GPT-3.5-Turbo to extract the answer.

\subsection{Results and Discussion}
The results are in Table~\ref{tab:results}.
In general, GPT-4o performs the best under all settings for LocateBench.  Gemini-1.5-pro is the second-best-performing model on LocateBench despite being the best-performing model on Pointing QA. Claude-3 Opus and Llava-1.6 lag behing on both tasks.


Overall, multi-choice by alphabet letters led to the highest accuracies.
Gemini-1.5-pro is most sensitive to prompt methods, showing a difference in performance between the best and worst settings of 43.9\%.
Claude-3 Opus is the least sensitive model, with a difference of 16.8\%. 
We plot Venn diagrams for model mistakes in Figures \ref{fig:venn-by-model} and \ref{fig:venn-by-prompt}.

The accuracy of GPT-4o still greatly lags behind the human accuracy of 95\%. Current proprietary LLMs still have room for improvement when it comes to object locating.
We include some hard examples where all models fail in Figure~\ref{fig:hard-examples}.

\begin{figure}[t]
    \centering
    \begin{subfigure}[t]{0.22\textwidth}
            \centering
            \includegraphics[width=\linewidth]{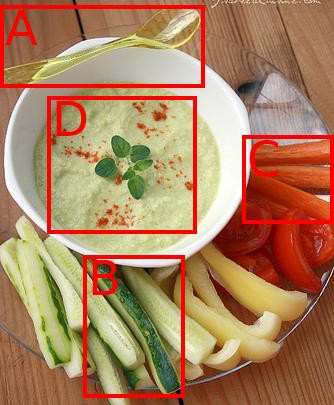}
            \caption{Which food is orange and is very good for you?}
            \label{fig:7w-od}
    \end{subfigure}
    \hspace{0.01\textwidth}
        \begin{subfigure}[t]{0.22\textwidth}
            \centering
            \includegraphics[width=\linewidth]{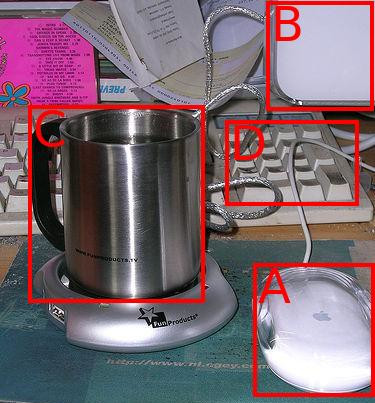}
            \caption{Which box frames the white?}
            \label{fig:7w-aa}
    \end{subfigure}
    \begin{subfigure}[t]{0.22\textwidth}
            \centering
            \includegraphics[width=\linewidth]{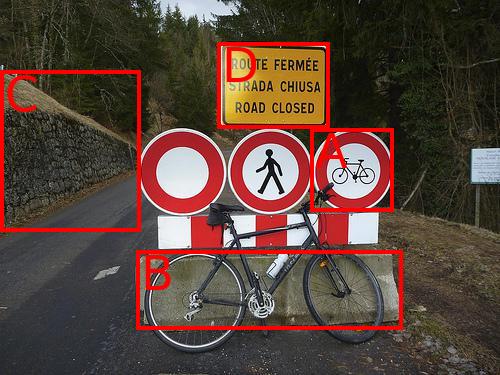}
            \caption{Which item is the fastest in the image?}
            \label{fig:7w-aq}
    \end{subfigure}
    \hspace{0.01\textwidth}
    \begin{subfigure}[t]{0.22\textwidth}
            \centering
            \includegraphics[width=\linewidth]{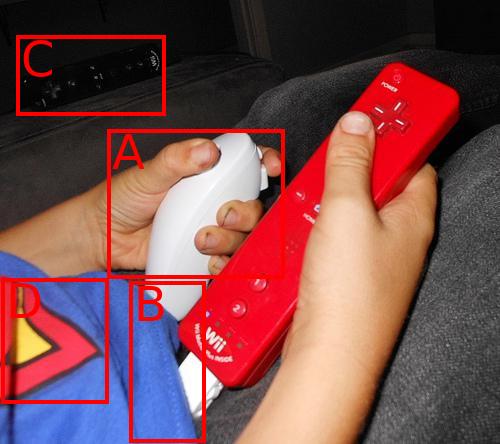}
            \caption{Which hand has the red remote?}
            \label{fig:7w-na}
    \end{subfigure}

    \caption{
    Less ideal examples in Pointing QA (\S\ref{sec:pointing-qa}). 
    }
    \label{fig:example-7w}    
\end{figure}

\section{Comparison with Pointing QA}
\label{sec:pointing-qa}

Although Pointing QA~\citep{zhu2016cvpr} has the same objective as our LocateBench dataset, we argue that our benchmark dataset is necessary in the following aspects:

\paragraph{Candidate bounding boxes.} 99.8\% of the examples in Pointing QA contain bounding boxes that are either significantly overlapping with other boxes, or too small. Only 139 out of 57,265 total test examples remain after the filtering process specified in \S\ref{sec:construct}.

\paragraph{Noisiness.}
We manually inspect the remaining examples and find that about 20.9\% (29 out of 139) of the examples are noisy. Specifically, 14 examples have more than one acceptable answer (e.g., Figure~\ref{fig:7w-aa}), 7 examples have ambiguous questions (e.g., Figure~\ref{fig:7w-aq}), 7 examples have no correct candidate (e.g., Figure~\ref{fig:7w-na}) and 1 example's annotated answer is incorrect. LocateBench, on the other hand, has minimal noise due to the rigorous validation process done by the authors.

\paragraph{Complexity.}
We also find that 40 of the other 97 examples (41\%) do not require sophisticated interactions between the text and vision domains. For example, the question in Figure~\ref{fig:7w-od} is simplified by the fact that only one of the four candidate answer objects is orange.

In comparison, questions in LocateBench are more complicated in general.
We manually inspect 100 randomly sampled examples.
Only 13 can be solved without sophisticated interactions between the text and vision domain.
The descriptions of the other examples are either more fine-grained or about relative size, counting, or location relative to other objects in the image (Figure~\ref{fig:example}).

\section{Related Work}

\subsection{Benchmarking Vision Language Models}
\label{sec:related-datasets}

In addition to the Pointing QA dataset from Visual7W~\cite {zhu2016cvpr}, recent benchmarks that involve explicit visual reference include the VCR~\citep{zellers2019vcr} and Pointer QA~\citep{mani2020point} datasets, which require models to reason about a specified point in an image.  
Other benchmarks evaluate VLM capabilities more generally~\citep{hendrycks2021measuring,zhang2023m3exam,fu2023mme,yu2024mm,fu2024isobench}.

\subsection{Grouding Vision Language Models}
\label{sec:related-models}

There have many works aimed at equipping VLMs with the ability to reference and ground objects in images \citep{lisa,lisap,zhao2023bubogpt,wang2023allseeing,wang2023visionllm,pi-etal-2023-detgpt,chen2023shikra,xu2023pixel,peng2024grounding,you2024ferret,zhang2024ferret,glamm,zhang2024gptroi}.
They extend VLMs to enable them to take in regions of an image specified by segmentation masks as a part of their input, and to generate segmentation masks as part of their output.
Most of these models are based on existing pre-trained models, such as CLIP-ViT-L~\citep{vlt},  ViT-H SAM~\citep{vith}, Vicuna~\citep{vicuna2023}, LLaVA~\cite{liu2023visual}, and Alpaca~\cite{alpaca}.
They extend the backbone models and conduct further instruction tuning. 
We discuss the source of the instruction-tuning data in \S{\ref{sec:related-it-data}}.

\section{Conclusion}

In this work, we propose a new benchmark, LocateBench, which evaluates VLMs' ability to locate objects specified by natural language descriptions.
We experiment with a set of advanced proprietary models and with a diverse set of prompting methods, and we show that even the best model still significantly lags behind human performance.
Our work provides an easy-to-use, high-quality playground for future VLM developers looking to test their models' locating ability and improve the interpretability of the model performance, as the performance on LocateBench dissects the behavior on downstream tasks.

\section*{Limitations}


In this work, we focus on multi-choice problems with only four candidates.
This may not fully reflect the complexity of some real-world tasks.
We leave more challenging setups for future work.

Additionally, due to budget constraints, we make a few design choices when constructing the benchmark.
For example, we only use a single LLM (Reka Core) to convert descriptions to English questions.
Besides, our dataset is based on a compilation of existing datasets.
This follows the common practice of repurposing existing resources for LLM evaluation. 
For example, HotpotQA~\citep{yang-etal-2018-hotpotqa} and StrategyQA~\citep{geva-etal-2021-aristotle} are based on Wikipedia articles. 
Just like how VLMs may have been exposed to COCO data, many LLMs have been exposed to Wikipedia in their training data. 
Critically, these datasets' challenge comes in its addition of questions on top of Wikipedia. 
Analogously, we contribute questions and bounding-box-to-label mappings that are not in VLM training data. It is evident that our aforementioned contributions make for a real challenge to VLMs (even if they’ve been exposed to COCO data) as there is still a sizable gap between best VLM and human performance.



\bibliography{custom}

\appendix

\section{Related Work: Datasets for Instruction Tuning}
\label{sec:related-it-data}
Most works derive instruction-tuning data from existing datasets.
For example, \citet{lisa} utilizes image segmentation datasets such as ADE20K~\citep{ade20k}, COCO-stuff~\citep{coco_stuff}, LVIS~\citep{lvis} and referring expression datasets such as RefCOCO~\cite{refcoco}, RefCOCO+~\citep{refcocop}, RefCOCO-g~\citep{refcocog} and convert them into question-answer pairs with templates.
LISA++~\cite{lisap} further utilizes GPT-4v to generate question-answer pairs where the answer refers to multiple objects in the images. 
\citet{you2024ferret} propose the GRIT dataset, which combines the RefCOCO series datasets, Visual Genome~\cite{krishna2017visual}, Object365~\cite{obj365}, Flickr30k~\cite{flickr} and dialogue data generated with ChatGPT/GPT-4. 
\citet{peng2024grounding} propose GrIT composed with COYO-700M~\cite{coyo} and Lion-5b~\cite{schuhmann2022laionb}.

\section{Dataset License}

We use the following dataset in our work:

\begin{itemize}
    \item COCO~\citep[Common Objects in Context,][]{mscoco}: Available at \url{https://cocodataset.org/} under Creative Commons Attribution 4.0 License
    \item RefCOCO~\citep{refcoco} and RefCOCO+~\citep{refcocop}: Available at \url{https://github.com/lichengunc/refer}.
    \item RefCOCO-g~\citep{refcocog}: Available at \url{https://github.com/mjhucla/Google_Refexp_toolbox} under Creative Commons Attribution 4.0 International License.
    \item Visual7W~\citep{zhu2016cvpr}: Available at \url{https://ai.stanford.edu/~yukez/visual7w/}
    
\end{itemize}

\section{Instructions for Step 2 in \S{\ref{sec:construct}}}

Please check whether the description of the object applies only to the target candidate. If not, please edit the description.

\section{Experimental Details for LLaVA}

\begin{itemize}
    \item Model: \texttt{llava-hf/llava-v1.6-vicuna-7b-hf} and \texttt{llava-hf/llava-v1.6-mistral-7b-hf}.
    \item Hardware: NVIDIA A6000
    \item Library: We use transformers 4.42.0.
\end{itemize}


\begin{figure*}
    \centering
    \begin{subfigure}[t]{0.24\textwidth}
        \includegraphics[width=\textwidth]{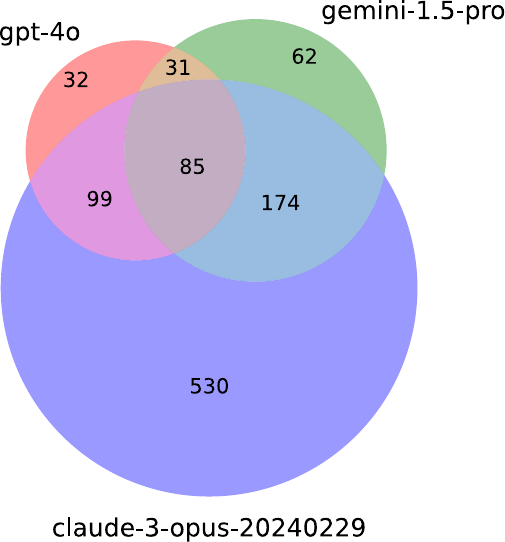}
        \caption{ABCD}
    \end{subfigure}
    \begin{subfigure}[t]{0.24\textwidth}
        \includegraphics[width=\textwidth]{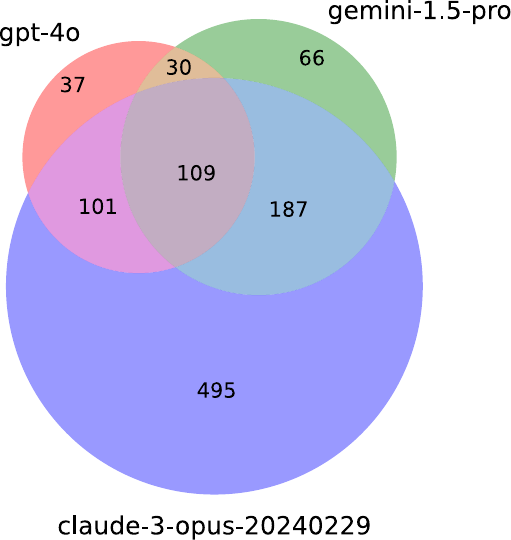}
        \caption{Color}
    \end{subfigure}
    \begin{subfigure}[t]{0.24\textwidth}
        \includegraphics[width=\textwidth]{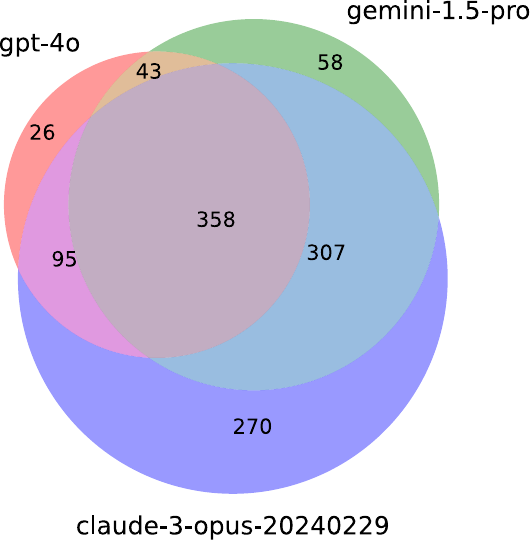}
        \caption{1-by-1}
    \end{subfigure}
    \begin{subfigure}[t]{0.24\textwidth}
        \includegraphics[width=\textwidth]{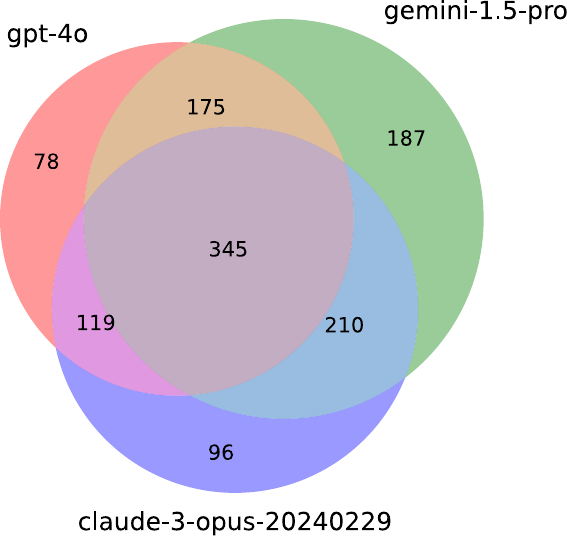}
        \caption{Coordinate}
    \end{subfigure}
    
    \begin{subfigure}[t]{0.24\textwidth}
        \includegraphics[width=\textwidth]{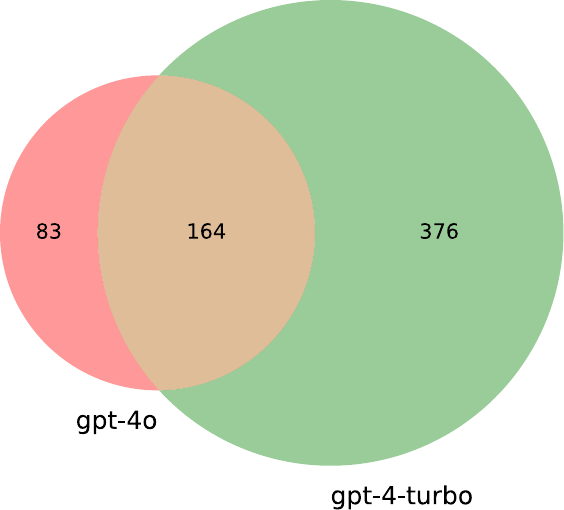}
        \caption{GPT-4 family (ABCD)}
    \end{subfigure}
    \begin{subfigure}[t]{0.24\textwidth}
        \includegraphics[width=\textwidth]{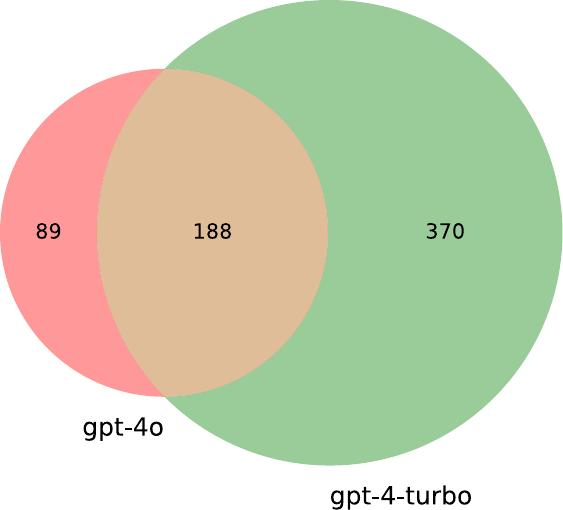}
        \caption{GPT-4 family (Color)}
    \end{subfigure}    
    \begin{subfigure}[t]{0.24\textwidth}
        \includegraphics[width=\textwidth]{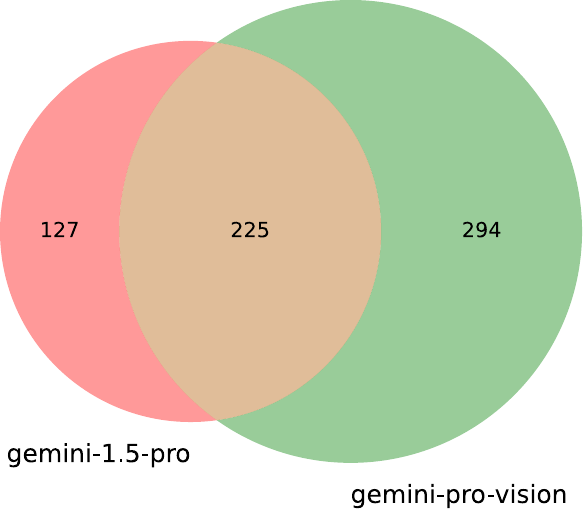}
        \caption{Gemini family (ABCD)}
    \end{subfigure}
    \begin{subfigure}[t]{0.24\textwidth}
        \includegraphics[width=\textwidth]{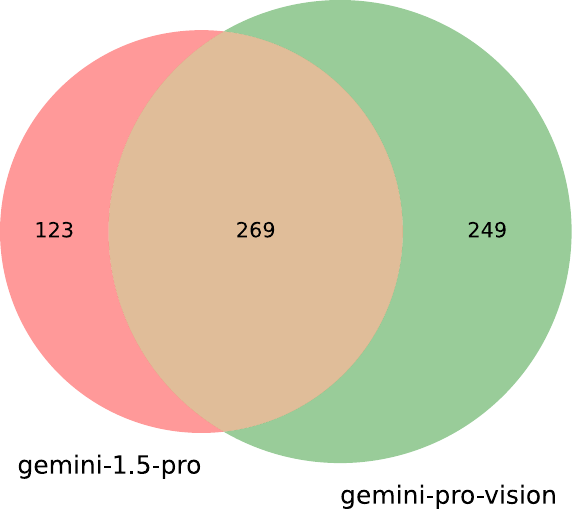}
        \caption{Gemini family (Color)}
    \end{subfigure}
    \caption{The Venn diagrams of the errors made by the three VLMs with different prompts.}
    \label{fig:venn-by-model}
\end{figure*}

\begin{figure*}
    \centering
    \begin{subfigure}[t]{0.30\textwidth}
        \includegraphics[width=\textwidth]{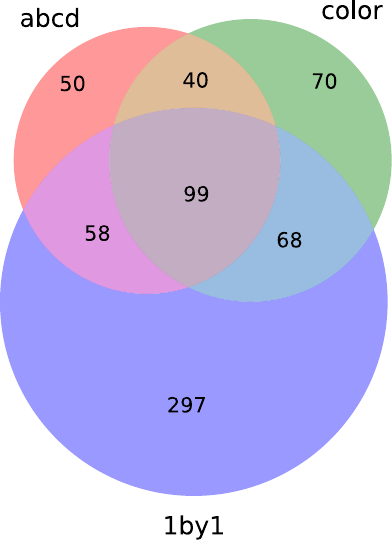}
        \caption{GPT-4o}
    \end{subfigure}
    \begin{subfigure}[t]{0.30\textwidth}
        \includegraphics[width=\textwidth]{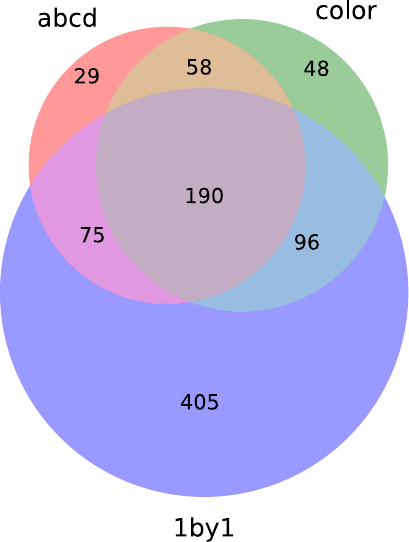}
        \caption{Gemini-1.5-pro}
    \end{subfigure}
    \begin{subfigure}[t]{0.30\textwidth}
        \includegraphics[width=\textwidth]{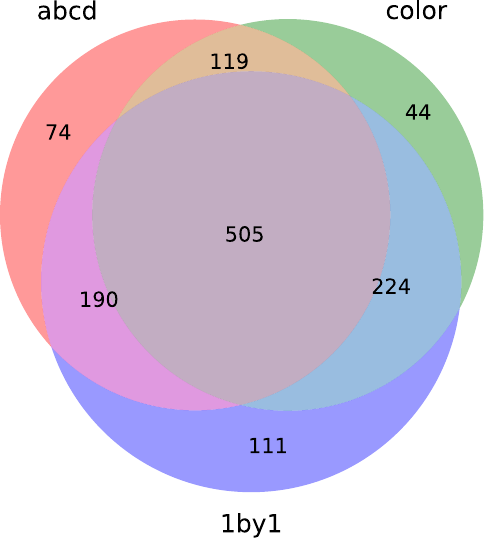}
        \caption{Claude}
    \end{subfigure}
    \caption{The Venn diagrams of the errors made by the three VLMs with different prompts.}
    \label{fig:venn-by-prompt}
\end{figure*}

\begin{figure*}[t]
    \centering
    \begin{subfigure}[t]{0.23\textwidth}
            \centering
            \includegraphics[width=\linewidth]{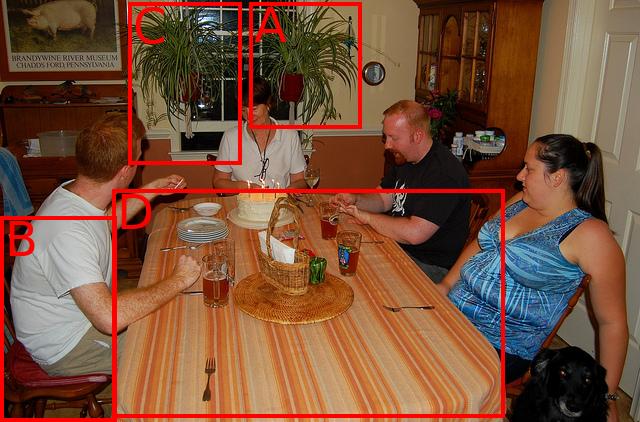}
            \caption{Which one contains a plant hanging next to a painting of a pig?}
    \end{subfigure}
    \hspace{0.01\textwidth}
    \begin{subfigure}[t]{0.23\textwidth}
            \centering
            \includegraphics[width=\linewidth]{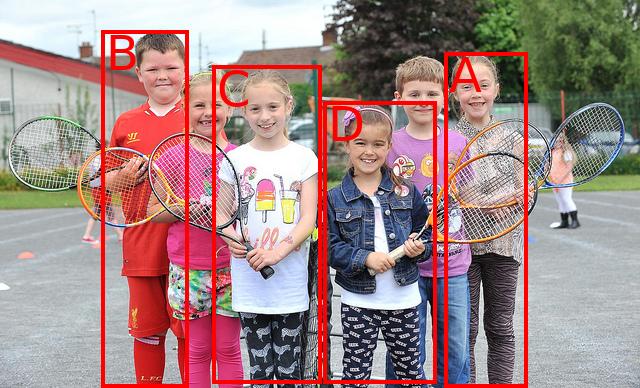}
            \caption{Which one contains the girl holding a blue racket?}
    \end{subfigure}
    \hspace{0.01\textwidth} 
    \begin{subfigure}[t]{0.23\textwidth}
            \centering
            \includegraphics[width=\linewidth]{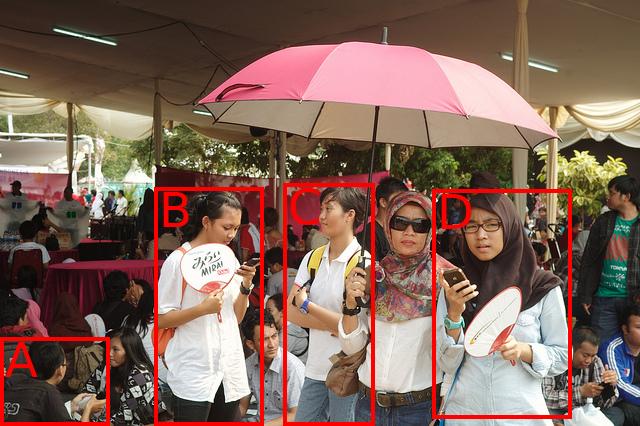}
            \caption{Which one contains a person with her fan over her chin?}
    \end{subfigure}
    \hspace{0.01\textwidth} 
        \begin{subfigure}[t]{0.23\textwidth}
            \centering
            \includegraphics[width=\linewidth]{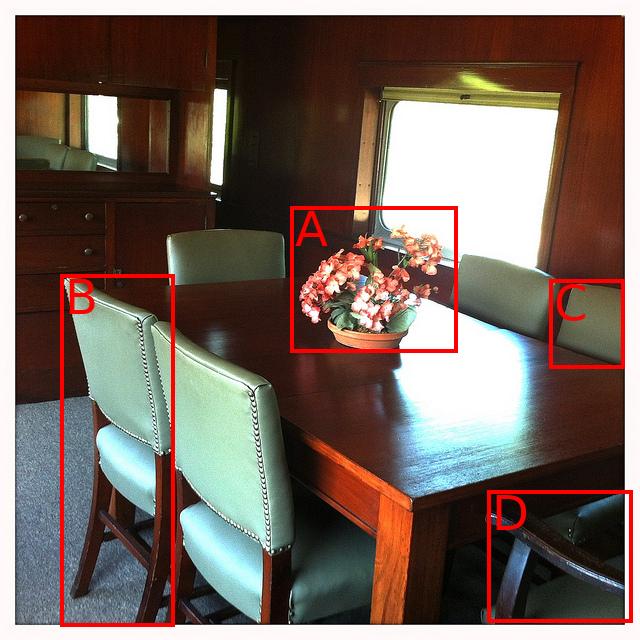}
            \caption{Which one contains the chair on the side away from the window?}
    \end{subfigure}

    \begin{subfigure}[t]{0.23\textwidth}
        \centering
        \includegraphics[width=\linewidth]{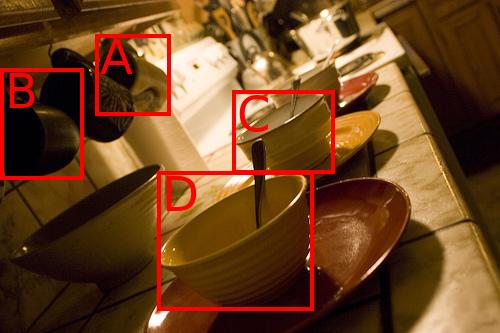}
        \caption{Which one contains a bowl with spoon sitting on a yellow plate?}
    \end{subfigure}
    \hspace{0.01\textwidth} 
        \begin{subfigure}[t]{0.23\textwidth}
        \centering
        \includegraphics[width=\linewidth]{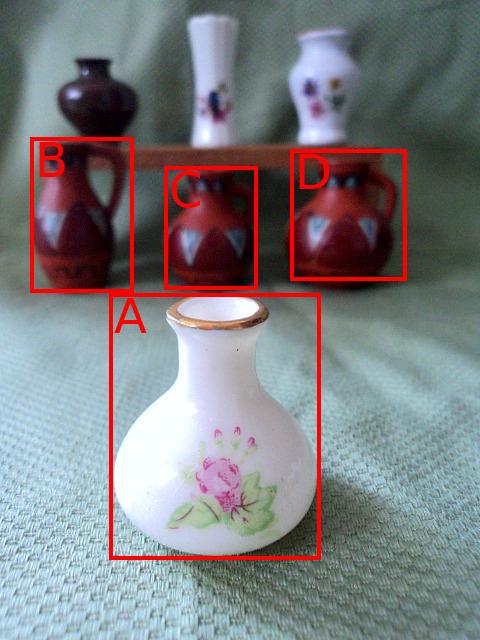}
        \caption{Which one contains the brown jar below the shortest white vase on the shelf?}
    \end{subfigure}
    \hspace{0.01\textwidth} 
        \begin{subfigure}[t]{0.23\textwidth}
        \centering
        \includegraphics[width=\linewidth]{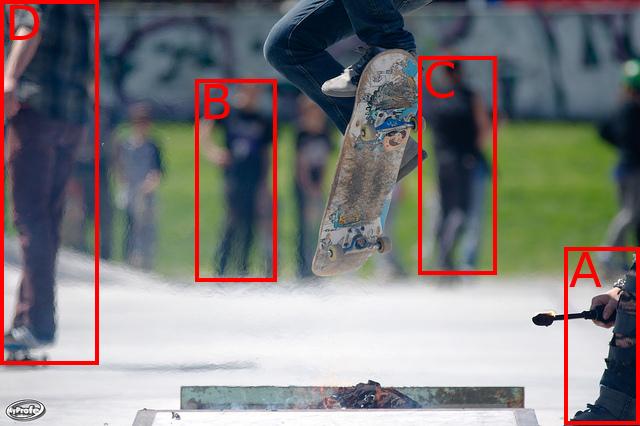}
        \caption{Which one contains a blurry person on a skateboard?}
    \end{subfigure}
    \hspace{0.01\textwidth} 
        \begin{subfigure}[t]{0.23\textwidth}
        \centering
        \includegraphics[width=\linewidth]{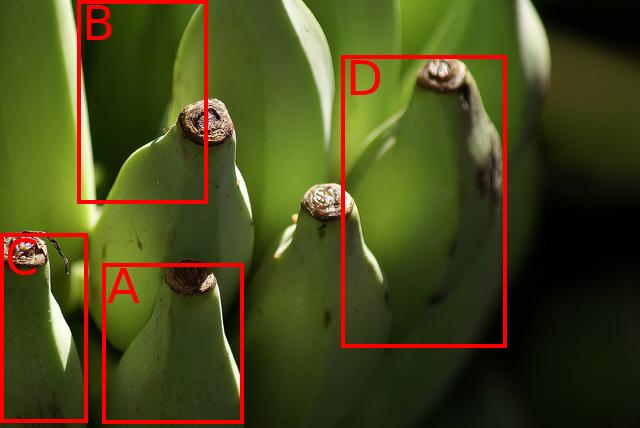}
        \caption{Which one contains the tip of banana closest to the corner?}
    \end{subfigure}
    
    \begin{subfigure}[t]{0.23\textwidth}
        \centering
        \includegraphics[width=\linewidth]{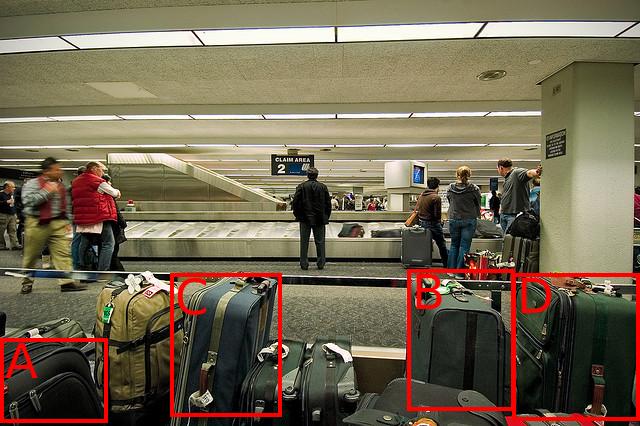}
        \caption{Which one contains the suitcase that is next to a standing green one?}
    \end{subfigure}    
    \hspace{0.01\textwidth} 
    \begin{subfigure}[t]{0.23\textwidth}
        \centering
        \includegraphics[width=\linewidth]{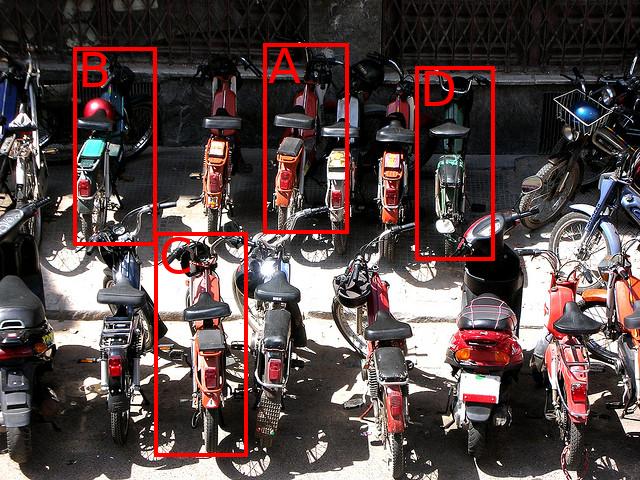}
        \caption{Which one contains the green bike to the left of the bike with the blue ball in its basket?}
    \end{subfigure}    
    \hspace{0.01\textwidth} 
    \begin{subfigure}[t]{0.23\textwidth}
        \centering
        \includegraphics[width=\linewidth]{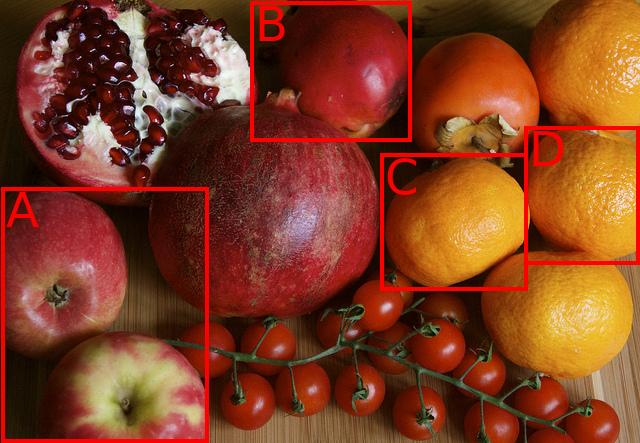}
        \caption{Which one contains the apple that is to the right of a pom?}
    \end{subfigure}    
    \hspace{0.01\textwidth}   
    \begin{subfigure}[t]{0.23\textwidth}
        \centering
        \includegraphics[width=\linewidth]{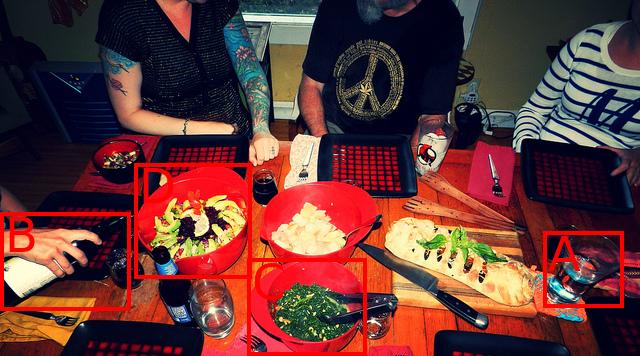}
        \caption{Which one contains the big red bowl with avocado in it?}
    \end{subfigure}   

    \begin{subfigure}[t]{0.23\textwidth}
        \centering
        \includegraphics[width=\linewidth]{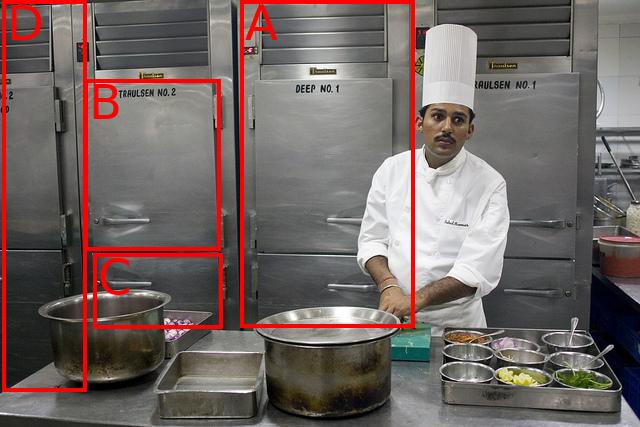}
        \caption{Which one contains the silver kitchen compartment with two lines of words on the door?}
    \end{subfigure}   
    \hspace{0.01\textwidth}   
    \begin{subfigure}[t]{0.23\textwidth}
        \centering
        \includegraphics[width=\linewidth]{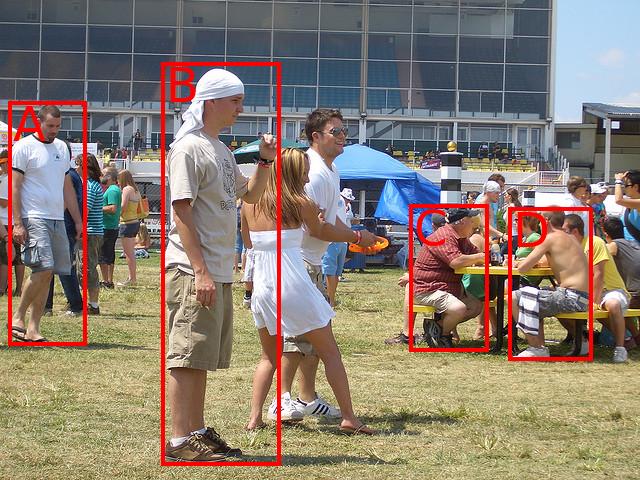}
        \caption{Which one contains the guy in white next to the person wearing a striped shirt?}
    \end{subfigure}   
    \hspace{0.01\textwidth}   
    \begin{subfigure}[t]{0.23\textwidth}
        \centering
        \includegraphics[width=\linewidth]{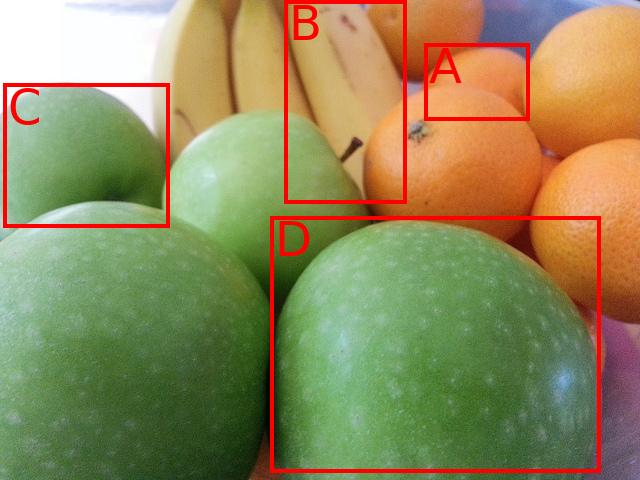}
        \caption{Which one contains an apple that is behind three other fruits?}
    \end{subfigure} 
    \hspace{0.01\textwidth}   
    \begin{subfigure}[t]{0.23\textwidth}
        \centering
        \includegraphics[width=\linewidth]{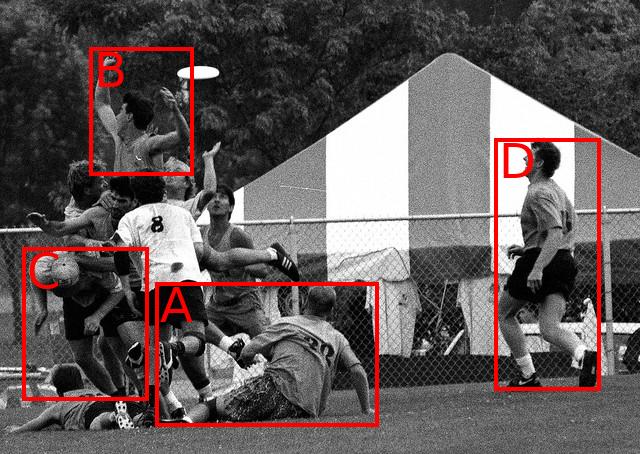}
        \caption{Which one contains the man who bends down?}
    \end{subfigure}
    \caption{Hard examples that all models got wrong in the multi-choice by alphabet letters (ABCD) setting.}
    \label{fig:hard-examples}    
\end{figure*}

\end{document}